\documentclass[10pt, a4paper]{article}

\usepackage[]{lrec2026} 

\title{Renaissance: A Framework for Investigating the Pretraining of Vision-Language Encoders}

\name{Clayton Fields, Casey Kennington} 

\address{Department of Computer Science\\
Boise State University \\
         \{claytonfields, caseykennington\}@boisestate.edu\\}

\abstract{
In the past several years there has been an explosion of available models for vision-language (VL) tasks. Unfortunately, the literature still leaves open a number of questions related to best practices in designing and training such models. Additionally, the limited programming tools available for  modeling make conducting VL research more difficult than necessary. In this paper, we seek to answer several questions related to the pretraining of VL encoders through meta-analysis. To conduct these experiments, we introduce a VL evaluation framework called Renaissance. In our first set of experiments, we show that we can save significant compute at little to no cost to downstream performance, by freezing large parts of VL models during pretraining.  In our second set of experiments, we examine the effect of basing a VL transformer on a vision model versus a text model. Renaissance offers a great deal of flexibility in creating, training and evaluating transformer encoders for VL modeling. Its source code will be made publicly available upon publication.
 \\ \newline  }

\begin{document}

\maketitleabstract

\section{Introduction}

In the span of a few years, dozens of vision-language (VL) transformers have appeared in the literature with a bewildering array of architectures and training methods (see \citet{fields2023vision} for a review). VL tasks, such as NLVR2 \citep{suhr2018corpus} where the model is tasked with answering questions about images require models to somehow represent and fuse both text and image information. Over the past few years, research in this field has been dominated by the study of generative, decoder-based models. Though generative models offer exciting possibilities and are required for tasks such as image captioning, encoder models are still well suited to classification tasks and are generally smaller and require significantly less compute for training and inference. These characteristics make encoder models a valuable asset for practical applications and for researchers who don't have access to large-scale compute resources. 
Unfortunately, as the focus of most research has shifted to very large, decoder-based models, the study of VL encoders has been neglected. This stands in contrast to the natural language procesing (NLP) domain, where studies such as \citet{rogers2021primer} have thoroughly investigated the inner workings and best training practices for NLP transformer encoders. To date, there have been only a handful of studies analyzing VL-transformer encoders, such as \citet{bugliarello2021multimodal}, and the collected literature still fails to address some very basic questions concerning VL modeling with transformers.



In this paper, we begin to address this gap by providing a systematic analysis geared toward shedding light on some basic aspects of training transformer encoders for VL modeling. As our focus is on the investigation of encoder models, we do not address generative tasks like image captioning here. In our first set of experiments (Section~\ref{sec:experiment-2}), we ask whether it is possible to minimize compute requirements by freezing parts of the model during pretraining and examining the effect on downstream performance. In our second and final set of experiments (Section~\ref{sec:experiment-3}) we compare the performance of a VL transformer based on a pretrained text encoder versus one based on a pretrained vision transformer. Both sets of experiments help establish best training practices for those interested in training VL transformers and hopefully also provide theoretical insight. To perform our experiments, we created a novel VL framework that we call \textit{Renaissance} that streamlines the ability to evaluate different VL model types (e.g., \textit{one-tower} and \textit{two-tower}, explained further in Section~\ref{subsubsec:renaissance:capabilities:model-types}) against a suite of benchmarks. This program offers a variety of training options and we hope that it proves a valuable tool for researchers. 

The specific contributions of this paper can be summarized as follows:
\begin{itemize}
    \item We introduce a software framework \textit{Renaissance} that offers a range of options for creating, training and testing VL transformer encoder models.
    

    \item We demonstrate that a great deal of compute can be saved by freezing parts of two-tower encoder models during pretraining. In particular, freezing the visual module can actually lead to small increases in performance. When both modules are frozen there is some loss in downstream performance, though the benefits may outweigh the costs for those with compute-limited training setups.

    \item We show that when training a one-tower encoder model, it is best to initialize the model's weights randomly than to use pretrained weights from either a text or a vision encoder model.
    
\end{itemize}








\section{Related Work}
\label{sec:related-work}

\subsection{Pretraining Vision-Language Transformers}


While the literature is now replete with VL models, the analysis of their performance and the establishment of best practices has mostly been applied to large decoder-based, generative models. Some examples include \citet{aghajanyan2023scaling}, \citet{mckinzie2024mm1} and \citet{tong2024cambrian}. While the studies offer a great deal of insight into generative models, those interested in using light-weight encoder models to achieve their goals are left with a number of questions unanswered. Some work has been devoted to the analysis of transformer encoders. \citet{bugliarello2021multimodal} examined pretraining of VL model and \citet{bugliarello2023measuring} provides an analysis of several models on what they term ``fine-grained" tasks. \citet{frank2021vision} examined the extent to which the vision and language modalities are actually integrated in VL transformers. As valuable as these studies have been, basic questions about VL transformers and their training remain open. These include: which architectures perform best in which situations, best practices for pretraining and what effects pretrained NLP and CV models have on downstream VL performance. This study begins to address some of these.

\subsection{Vision-Language Modeling Framework}

When using language models, researchers have a range of available software options that abstract many of the most difficult elements away from users. The most prominent example of this is the Huggingface model hub that specializes transformers. The Huggingface hub has many different model types devoted to NLP and several to computer vision (CV). Unfortunately, there are only a few VL models available and they don't easily lend themselves to the modifications that research often demands. In addition to the Huggingface Hub, there have been some efforts toward creating software frameworks primarily dedicated to multimodal modeling. LAVIS, introduced in \citet{li-etal-2023-lavis} by Salesforce, is one such platform. Though well programmed and relatively straight forward to use, this program supports relatively few VL models. To date, there is still a great deal of space for additional programming options in the field of vision language modeling, particularly for purpose of conducting research. To help fill this space, we created the Renaissance framework for VL modeling that we introduce in the next section.

\section{Renaissance: A Versatile Vision-Language Modeling Framework}
\label{sec:renaissance}

In this section, we describe the Renaissance framework that we use to complete all of the experiments in this study. Because this is its first introduction, we provide an extensive description of the framework and its capabilities.  We also introduce the pretraining tasks, finetuning tasks and the architectural elements required to understand the experimental procedures. 

\subsection{Functionality}
\label{subsec:renaissance:capabilities}
In this section we describe the functionality and various options available from the Renaissance framework. The most salient feature of the framework is its ability to easily change the basic architectural features of multi-modal transformers, then train and test them. By simply editing a configuration file, a user can choose a pretrained text encoder or a pretrained vision encoder from the Huggingface hub to insert into the model. In addition to the various architectural options, there are also a number of pretraining and fine-tuning tasks and options available. 

\subsubsection{Model Types}
\label{subsubsec:renaissance:capabilities:model-types}
\paragraph{One-Tower Encoder Modeling} A one-tower encoder model consists of an embedding layer, and a single transformer encoder module followed by a classification layer. Previous examples of one-tower encoders include models such as UNITER \citep{chen2020uniter} and VisualBERT \citep{li2019visualbert}. In principle, one-tower encoders are very much like NLP encoders such as BERT \citep{devlin2019bert} and ELECTRA \citep{clark2020electra} with some adaptations for the vision language domain.

\begin{figure}[t]
\centering
\includegraphics[scale=0.2]{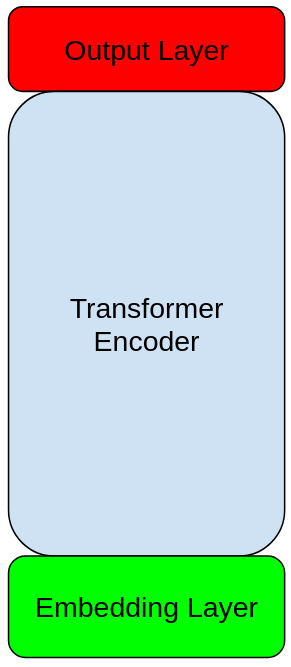}
\caption{A visual representation of a one-tower vision-language encoder model.}
\label{fig:one-tower}
\end{figure}

One of the key adaptations is that a VL model's embedding layer must accommodate both textual and visual features. For NLP models such as BERT, the embedding layer consists of a single large matrix where each column in the matrix is a vector representing the words in the model's vocabulary. While one-tower VL encoders also have this feature, they have additional components that can process an image into a sequence of vectors. In the current version of the program, the embedding layer will always consist of BERT-style word-piece embeddings for text \citep{devlin2019bert} and patch embeddings for image features. Patch embeddings were first introduced as part of the ViT model in \citet{dosovitskiy2020image}. Here an image is split in small square patches, the patch is then flattened into a vector and projected to the correct embedding dimension. 
In future versions of the program we hope to include support for using grid features derived from a convolutional neural network to embed images \citep{huang2020pixel}.

The second major component of one-tower encoder models is the transformer encoder stack. The encoder stack for VL models is architecturally the same as those found in NLP transformers. Here, the only major difference is that encoder's weights are derived from training on VL tasks. Renaissance supports the use of most text models on the hub as encoder modules and select variety of vision transformer models on the hub. Specifically, vision models based on the transformer \citep{vaswani2017attention}, such as ViT \citep{dosovitskiy2020image}, DeiT \citep{touvron2021training}, DINO \citep{caron2021emerging} or BeIT \citep{bao2021beit}. Convolutional models such as ResNet \citep{he2015deepresiduallearningimage}, and hybrid models such as ConvNeXT \citep{liu2022convnet2020s} are not supported for one-tower modeling. Finally, the classification layer is no different than those found in any other deep learning model. They consist of one or two linear layers that output a score for each possible outcome in the target distribution.

\begin{figure}[t]
\centering
\includegraphics[scale=0.25]{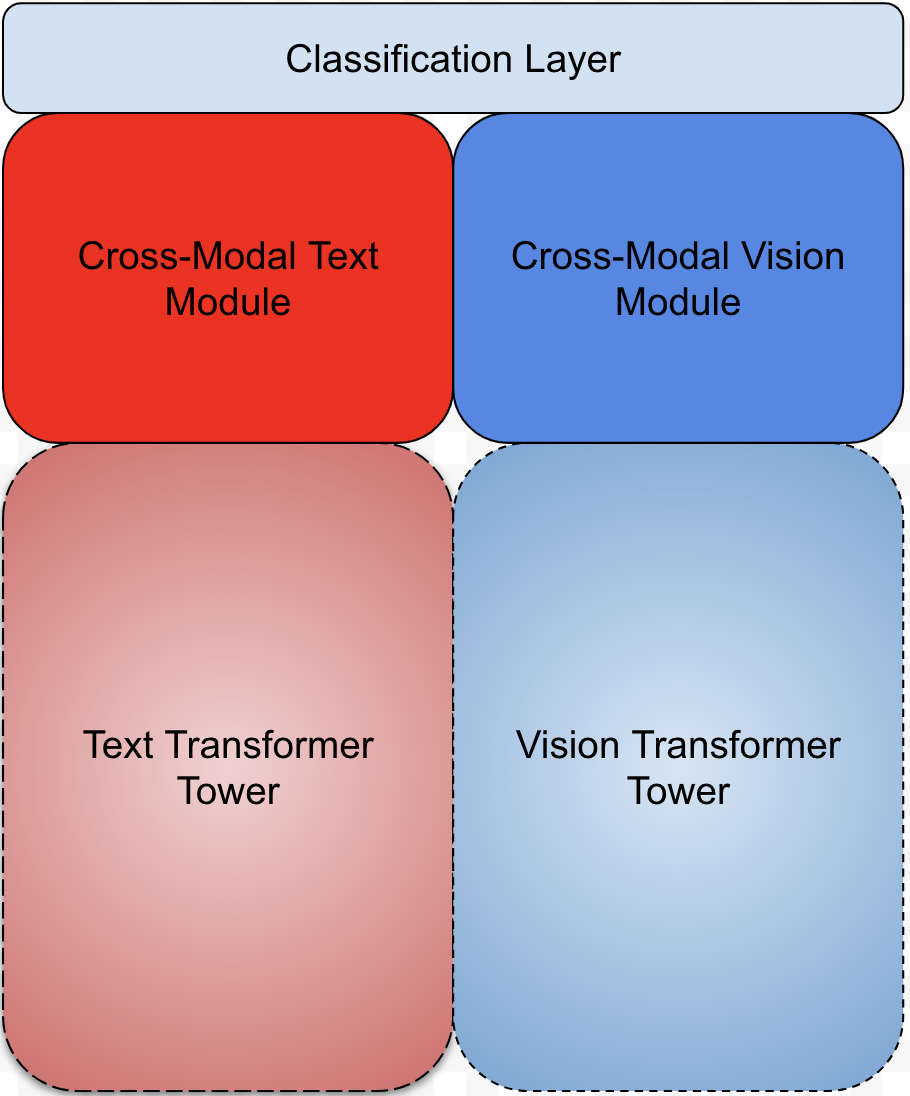}
\caption{A visual representation of a two-tower vision-language encoder model.}
\label{fig:two-tower}
\end{figure}


\paragraph{Two-Tower Encoder Modeling} A two-tower encoder model consists of a text-transformer model, a vision model and set of cross-modal layers that combine the output of  each model into a multimodal feature using cross-attention \citep{lu2019vilbert}. In cross-attention layers, the key and value vectors from the visual stream are passed to the multi-head attention mechanism of the textual stream. The key and value vectors from the text stream are also passed to the attention heads in the visual stream resulting in a multi-modal output. Figure~\ref{fig:two-tower} shows a simple visual representation of a two-tower model. Because the vision and text modules are separate and the vision and text streams only interact in the cross-modal layers. This is in contrast to one-tower models where visual and textual features interact throughout the model. Another distinction is that for two-tower models the visual and textual features need not be embedded in the same vector space because each encoder module is associated with its own embedding layer. Some previously introduced examples of two-tower transformers are METER \citep{dou2022empirical} and BridgeTower \citep{xu2023bridgetower}. 

In previously released two-tower encoders, the text encoder modules are architecturally much like BERT and the vision modules much like ViT \citep{dosovitskiy2020image}. Renaissance allows users to create new two-tower models with most of the vision modules on the hub as a vision module. A novel feature of this framework is that users can also incorporate convolutional models such as ResNet \citep{he2015deepresiduallearningimage} and hybrid models such as ConvNext \citep{liu2022convnet2020s}. Most text transformers on the hub can be used as text modules in two-tower models. The layers in the cross-modal module are based on the implementation from the LXMERT model \citep{tan2019lxmert}. Users can choose the dimension and number of cross-modal layers and the number of attention heads per layer. Finally, the classification layer is essentially the same as those found in one-tower models.  



\subsubsection{Training and Configuration Options}
\label{subsubsec:renaissance:capabilies:training}

Beyond providing flexibility in basic architecture design, the program also provides several options for training and configuring models. The most notable of these features are discussed in this subsection. 

\paragraph{Random Weight Initialization} Multi-modal models are often initialized with weights from pretrained text or image models. For instance VisualBERT is initialized with the weights from the text model BERT \citep{devlin2019bert} and ViLT with weights from the image transformer ViT \citep{dosovitskiy2020image}. When doing research it is often useful to initialize model weights randomly and train from scratch. This is useful for establishing baselines in experiments and as we show in Section \ref{sec:experiment-3}, can be beneficial to the performance of one-tower models. Users can randomly initialize encoder weights by simply changing settings in a configuration file. 

\paragraph{Manually Configure Model Dimensions} By default, the dimensions of encoder modules is determined by the huggingface hub. However, when model weights are set to be randomly initialized users can manually specify the dimensions of encoder modules. This allows users to easily create completely novel architectures. As an example, consider a one-tower encoder where the encoder is based on ELECTRA-Small. By default, ELECTRA has a hidden size of 256, an embedding size of 128, an intermediate size of 1028 and 12 layers. Any of these hyperparameters can be altered to create encoders of the desired shape and size.

\paragraph{Freeze Modules During Training} It is also easy to freeze the weights of any of the models modules during training. In addition to being useful for research purposes, this feature allows the user to significantly cut the compute costs of training. In practice, freezing the pretrained weights of a model's encoder module can be quite useful and is featured in our first set of experiments. 

\subsubsection{Pretraining Tasks}
\label{subsec:pretraining}
Currently, our program supports two pretraining tasks, masked language modeling and image-text matching. Models can be pretrained with either of these tasks individually or both in conjunction. Using both tasks in conjunction is a common approach found in the literature. Both tasks are briefly described in the list immediately below. A more thorough description can be found in \citet{fields2023vision}.
\begin{itemize}
    \item Masked language modeling (MLM) tasks the model with guessing a masked word based on the image features and the unmasked words. The MLM task was first introduced in \citet{devlin2019bert}. In the original task, the model's prediction is based only on the unmasked words in sequence of text. In the multimodal setting, the model's prediction is based on the unmasked words as well as the associated image features. 

    \item Image-text matching is a binary task where the model is presented with an image-text pair and must determine if the text actually describes the image. Positive pairs are simply the original pairings from the chosen datasets; for negative pairs a sentence is paired with a randomly chosen image from the dataset. This task is much like, and was inspired by, the next sentence prediction task that was also used in training BERT.
\end{itemize}

At the time of this writing, Renaissance currently supports four different multimodal datasets for pretraining. These four are Visual Genome \citep{krishna2017visual}, MSCOCO \citep{lin2014microsoft}, Conceptual Captions \citep{sharma-etal-2018-conceptual} and SBU Captions dataset \citep{ordonez2011im2text}. They can also be used by themselves or in any combination for pretraining models. These four datasets are commonly used in the pretraining of VL models. 

\subsubsection{Downstream Vision-Language Tasks}
\label{subsubsec:renaissance:capabilities:downstream}
In order to test and evaluate models, renaissance currently has five downstream VL tasks implemented. They are listed below with a brief description of each task. 
    \begin{enumerate}
        \item \textbf{NLVR2} NLVR2 stands for Natural Language Reasoning for Real and was introduced in \citet{suhr2018corpus}. Here a model will be given two images and must answer a true or false question about them. The addition of second image also makes this quite a challenging task. NLVR2 is very commonly used to benchmark VL models. 
        
        \item \textbf{SNLI-VE} In the SNLI-VE task is  a model is presented with an image text-pair and must determine if the image entails the sentence, contradicts the sentence or is neutral with respect to the sentence. It was introduced in \citep{xie2019visual}. This task tends to be less challenging than the previous and requires less time to fine-tune and evaluate. Though it appears less commonly in the literature, its quick training time makes it very useful as a model development tool.

        \item \textbf{Reference Resolution with RefCOCO} In this final task a model is presented with an image that is segmented into several objects and a sentence describing one of these objects. The model must then determine which object the sentence is referring to. The RefCOCO dataset was introduced in \citep{kazemzadeh2014referitgame}.
        
        \item \textbf{Multimodal Retrieval with MSCOCO and Flickr30k} Multimodal retrieval tasks correspond to activities such as an internet image search. Here a model is given a string of text and must rank a number of images according to how relevant they are to the sentence. The converse process, an image is provided and the model must rank a series of sentences, is also implemented. Our program supports fine-tuning and evaluating both retrieval tasks on the MSCOCO \citep{lin2014microsoft} and Flickr30k datasets. 
        
        \item \textbf{Visual Question Answering}
        In this task a model is presented with an image and a question pair and must choose the correct answer from a given set possible choices or generate a free-form answer. The visual question answering task (also called VQA) was introduced in \citet{antol2015vqa}. This task is a commonly used benchmark in the VL field and is also quite challenging. 

    \end{enumerate}

\begin{figure}[h]
\centering
\includegraphics[scale=0.8]{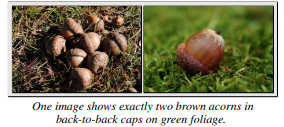}
\caption{An example from the NLVR2 dataset. Image from \citet{suhr2018corpus}.}
\label{fig:nlvr2}
\end{figure}

\section{Experiment 1: Freezing Encoder Modules During Pretraining}
\label{sec:experiment-2}

\subsection{Purpose} In our first set of experiments, we ask \emph{what is the effect of freezing the weights of various parts of the model during pretraining}? Specifically, if we initialize the vision and text modules of a two-tower encoder with pretrained models from their respective domains, can we freeze one or both of these modules during pretraining? Freezing both modules means that we would only be pretraining the cross-modal and output layers of the model. Pretraining is usually the most compute intensive aspect of model development and we can reduce the GPU memory use and the overall compute required by freezing parts of the model. The compute savings would allow researchers to pretrain models that might otherwise be too large for their hardware to support.\footnote{Alternatively, they might also train smaller models at higher batch sizes and possibly obtain better results, as was found in \cite{Fields2023-zq}.} 

Given that both vision and text encoder modules are pretrained in their respective domains, it makes intuitive sense that we might be able to skip at least some portion of their pretraining. These experiments should demonstrate empirically whether or not this is the case. Furthermore, the creators of the dual encoder\footnote{Dual encoder models are a simpler model type that is currently not available on Renaissance.} model LiT \citep{zhai2022lit} found that they obtained slightly better results from freezing the model's vision encoder. This experiment will also afford us the opportunity to see if a similar effect will hold for two-tower encoder models. 

\subsection{Experimental Setup and Procedure}
 
To begin, we use Renaissance to construct a set of two-tower models with ELECTRA-Small \citep{clark2020electra} as the text encoder and DeiT-Tiny \citep{touvron2021training} as the image encoder. Both of these models are both quite small and efficient, and we chose them to expedite the training process. We set the cross-modal encoder module of each model to contain two sets of six transformer layers  each with a hidden size of 256 and 4 attention heads. In total, we pretrain four model variations, a baseline with both modules unfrozen, one with the text encoder frozen, one with the image encoder frozen and one with both encoder modules frozen. Each model is pretrained for 100k steps at a batch size of 704 using the masked language modeling and image-text matching tasks described in Section~\ref{subsec:pretraining}. All models are trained using two NVIDIA L40s GPUs. We use two of the four pretraining datasets, MSCOCO and Visual Genome, which were described in the same section. Finally, we finetune and evaluate our models on three of the five VL tasks described in Section~\ref{subsubsec:renaissance:capabilities:downstream}: SNLI-VE, NLVR2 and reference resolution with RefCOCO. A crucial point to consider is that the no model weights are frozen during finetuning. 

\subsection{Results} The results for this experiment are summarized in Table~\ref{table:freeze-model}. We see that we can obtain similar results by freezing one or both of the previously trained encoder modules during pretraining with only mild ill effect. On the SNLI-VE task, the difference between training the whole model and freezing one or both modules is very slight indeed. When freezing the vision module, the downstream results for SNLI-VE are essentially identical and we see only a slight drop in performance compared to the baseline model. We see a slightly different pattern on the NLVR2 task, however. Here,the results for the baseline model and two models with a single encoder frozen have almost identical results. The model with the visual encoder frozen produces the best score on the reference resolution task. The baseline model and the model with both modules frozen preform very nearly identically, while the the model with the text encoder only frozen scores the worst. Of the four models, \emph{the overall best performance is achieved by freezing only the vision encoder}. 

This is a fairly remarkable result as we can significantly cut the GPU memory required during pretraining by freezing so many of the model's weights. This is especially significant because pretraining the model is by far the largest compute cost we see during training. We should also note that this effect is somewhat similar to a phenomenon noted in the training of the dual encoder LiT \citep{zhai2022lit} (dual encoders have two encoder stacks but lack the a cross-modal fusion module). Here \citeauthor{zhai2022lit} found that they can obtain better results from freezing the image encoder during training. Though our model architecture is different, we observe a somewhat similar effect. 


\begin{table*}
\centering
\begin{tabular}{|l|l|l|l|l|l|l|l|}
 \hline
 \textbf{Text Encoder} & \textbf{Vision Encoder}  & \textbf{SNLI-VE}  & \textbf{NLVR2} & \textbf{Ref. Res.} & \textbf{Avg.} \\ 
 \hline
 \hline
  Unfrozen & Unfrozen  & 0.741 & 0.672 & 0.724 & 0.712\\
 \hline
 Frozen & Unfrozen  & 0.735 & \textbf{0.675} & 0.702 & 0.704 \\
 \hline
 Unfrozen & Frozen  & 0.741 & 0.672 & 0.740 & \textbf{0.717}\\
 \hline
 Frozen & Frozen & 0.738 & 0.665 & 0.721 & 0.708 \\
 \hline
\end{tabular}
\caption{\label{table:freeze-model}
\textbf{Results for Freeze Module Study} 
} All models are trained for 100k steps at a batch size of 704, the largest size our server's could accomodate. All results are calculated on dev sets for each task.
\end{table*}

\section{Experiment 2: Text Encoder vs. Vision Encoder}
\label{sec:experiment-3}

\subsection{Purpose} In the previous set of experiments we focused on training two-tower models. In our final experiment, we examine the behavior of one-tower models (described in Section~\ref{subsubsec:renaissance:capabilities:model-types}). To date, most of these models have been derived from text encoder models such as BERT \citep{devlin2019bert}. A less explored approach is to base such models on transformer based vision models such as ViT \citep{dosovitskiy2020image}; this is the approach of the one-tower VL transformer called ViLT \citep{kim2021vilt}. In this experiment we ask if one strategy is superior to another when training and evaluating under otherwise similar conditions? More simply put, are one-tower encoders more effective when based on a vision encoder or a text encoder.  In addition to providing guidance to future practitioners of VL modeling, answering this question should also provide interesting results from purely theoretical as well as practical perspectives.

\subsection{Experimental Setup and Procedure} To make this experiment as fair a comparison as possible, we select a vision transformer and a text transformer model as close in size to each other and architecture as possible. Toward this end we used BERT \citep{devlin2019bert} as our text-encoder model and ViT \citep{dosovitskiy2020image} as our vision encoder model. The encoder towers in each of these models were consciously designed to have nearly identical dimensions with each encoder module containing 110M parameters. We employ patch embeddings for visual tokens and word-piece embeddings for visual tokens in all models. The resultant models will be close to identical, save that the weights of one are derived from vision pretraining and the other from text/language pretraining. As a baseline, we also train a randomly initialized version based on the BERT architecture. Finally, we train each model for 50k steps with a batch size of 512 using masked-language modeling and image-text matching. Again we use MSCOCO and Visual Genome as training datasets and evaluate on the 3 VL tasks described in the previous experiments.  

\subsection{Results}

The results for this experiment are displayed in Table \ref{table:text-vs-vision}. According to our analysis, there doesn't appear to be a significant advantage in basing a one-tower encoder model on either text or vision. Surprisingly, \emph{the randomly initialized model that we trained as a baseline scored the best on all three downstream tasks}. These are very much unexpected results. Though we didn't have an intuition as to whether text or vision would perform better, we didn't expect the downstream results to be so similar and to be inferior to a randomly initialized variation. These results are especially notable since one of the few in depth analyses of VL models, \citet{frank2021vision}, indicates that the interaction between the visual and language modalities are not symmetric. That study used probing techniques to show that VL transformers learn to use vision-for-language more than language-for-vision. Our best explanation of this phenomenon is that that one-tower models do not make use of the individual visual or textual modalities, but instead converge to values not dependent on either.

Another notable conclusion of this experiment and the preceding ones, is that two-tower models are in general much more parameter efficient than one-tower models. The one-tower models used in this experiment are relatively large, each containing more than 100M parameters. While the two-tower models in the previous experiments contain less than 40M parameters. Nonetheless, the two-tower models outperform those in this final experiment using the same datasets for training and evaluation. In previous studies, one-tower models such as ViLT \citep{kim2021vilt} that have similar architectures, have obtained significantly better results than those displayed here. They do so by using more data and enormous pretraining batch sizes that require significantly more compute than we used here. Though only a preliminary finding, this insight might prove valuable to those interested in efficient VL modeling.

\begin{table*}
\centering
\begin{tabular}{|l|l|l|l|l|l|l|}
 \hline
 \textbf{Encoder}  & \textbf{SNLI-VE}  & \textbf{NLVR2} & \textbf{Ref. Res.} & \textbf{Avg.}\\ 
 \hline
 \hline
Random   & \textbf{0.699} & \textbf{0.551} &  \textbf{0.554} & \textbf{0.601} \\
 \hline
 ViT  & 0.685 & 0.534 & 0.522 & 0.581\\
 \hline
 BERT  & 0.692 & 0.545 & 0.507 & 0.581\\
 \hline
\end{tabular}
\caption{\label{table:text-vs-vision}
\textbf{Preliminary Results for Text vs Vision Encoder Study} All models are trained for 100k steps at a batch size of 512. All results are calculated on dev sets for each task.
}
\end{table*}

\section{Future Directions}
\label{sec:future-directions}

As Renaissance evolves, we hope to incorporate a number of additional features that are not available in the current version. The capabilities that we plan to add are discussed below.  

\paragraph{Model Types}
There are several model types, beyond one-tower and two-tower encoders, that we plan to support in future versions. These include, dual encoder, encoder-decoder and decoder only model types (see \citet{fields2023vision} for explanations of and examples for each type). By virtue of adding these model types, we also hope to include the ability to generate text for tasks such as image captioning. 

\paragraph{Additional Tasks}
In addition to more model architectures, we hope also to add additional tasks for both pretraining and finetuning. Some asks we intend to add are contrastive learning, reference resolution and visual question answering as pretraining tasks. Further we also hope to add downstream tasks such as image captioning to give the a wider variety of settings to use and evaluate various model architecture.

\paragraph{Analysis of VL Transformers and Pretraining}
Because of the field of VL is rapidly evolving there are many possible future directions for research. We will mention a few. Though we have touched on some of the more basic aspects of training in this study, a systematic study of how each pretraining task contributes to downstream performance would be very illuminating, as would testing other tasks such as visual grounding or visual question answering in pretraining. A thorough investigation of which architectures are best used in which circumstances would also be a worthwhile endeavor.

\section{Conclusion}
\label{sec:conclusion}

In this study, we have examined some basic features of pretraining vision-language transformers. In addition to the experiments that we've performed, we also introduced a flexible vision-language modeling framework called Renaissance, the source code for which will by made available upon publication. In our first set of experiments we showed that pretrained vision and text modules can be frozen during vision-language pretraining with only small losses in downstream performance. This finding opens the possibility of training VL models whose size might normally exceed one's compute budget. In our second and final experiment we compared of effect of basing a one-tower encoder model on a text transformer versus a vision transformer. Surprisingly, our results indicate that neither strategy is superior to the other and that randomly initializing model weights yields the best results. We therefore recommend training one-tower models from scratch when possible. We conclude this study with the observation that multi-modal modeling is a rapidly expanding pursuit and we hope that this paper is among the first of many that aim to shed light on this dynamic and exciting field of deep learning. 

\section*{Ethics Statement}
The authors have no ethical conflicts to report. 







\section{Bibliographical References}\label{sec:reference}

\bibliographystyle{lrec2026-natbib}
\bibliography{lrec2026-example}


\end{document}